\documentclass[acmsmall]{acmart}
\usepackage{graphicx}
\usepackage{psfrag}
\usepackage{subfigure}
\usepackage{url}
\usepackage{stfloats}
\usepackage{amsmath}
\usepackage{epsfig}
\usepackage{algorithm, algorithmic}
\usepackage{color}
\usepackage{booktabs} 

\def\BibTeX{{\rm B\kern-.05em{\sc i\kern-.025em b}\kern-.08emT\kern-.1667em\lower.7ex\hbox{E}\kern-.125emX}}

\begin{document}
\title{Face Clustering for Connection Discovery from Event Images}

\author{Ming Cheung}
\orcid{0000-0003-4646-1980}
\affiliation{%
  \institution{Social Face Limited}
  \country{Hong Kong, China}
}
\email{ming@socialface.ai}


\renewcommand{\shortauthors}{M. Cheung}

\begin{abstract}
Social graphs are very useful for many applications, such as recommendations and community detections.
However, they are only accessible to big social network operators due to both data availability and privacy concerns.
Event images also capture the interactions among the participants, from which social connections can be discovered to form a social graph.
Unlike online social graphs, social connections carried by event images can be extracted without user inputs, and hence many social graph-based applications become possible, even without access to online social graphs.
This paper proposes a system to discover social connections from event images.
By utilizing the social information from even images, such as co-occurrence, a face clustering method is proposed and implemented, and connections can be discovered without the identity of the event participants.
By collecting over 40000 faces from over 3000 participants, it is shown that the faces can be well clustered with 80\% in F1 score, and social graphs can be constructed.
Utilizing offline event images may create a long-term impact on social network analytics.
\end{abstract}


\maketitle
\section{Introduction}
\begin{figure}
\centering 
\includegraphics[width=3.5in]{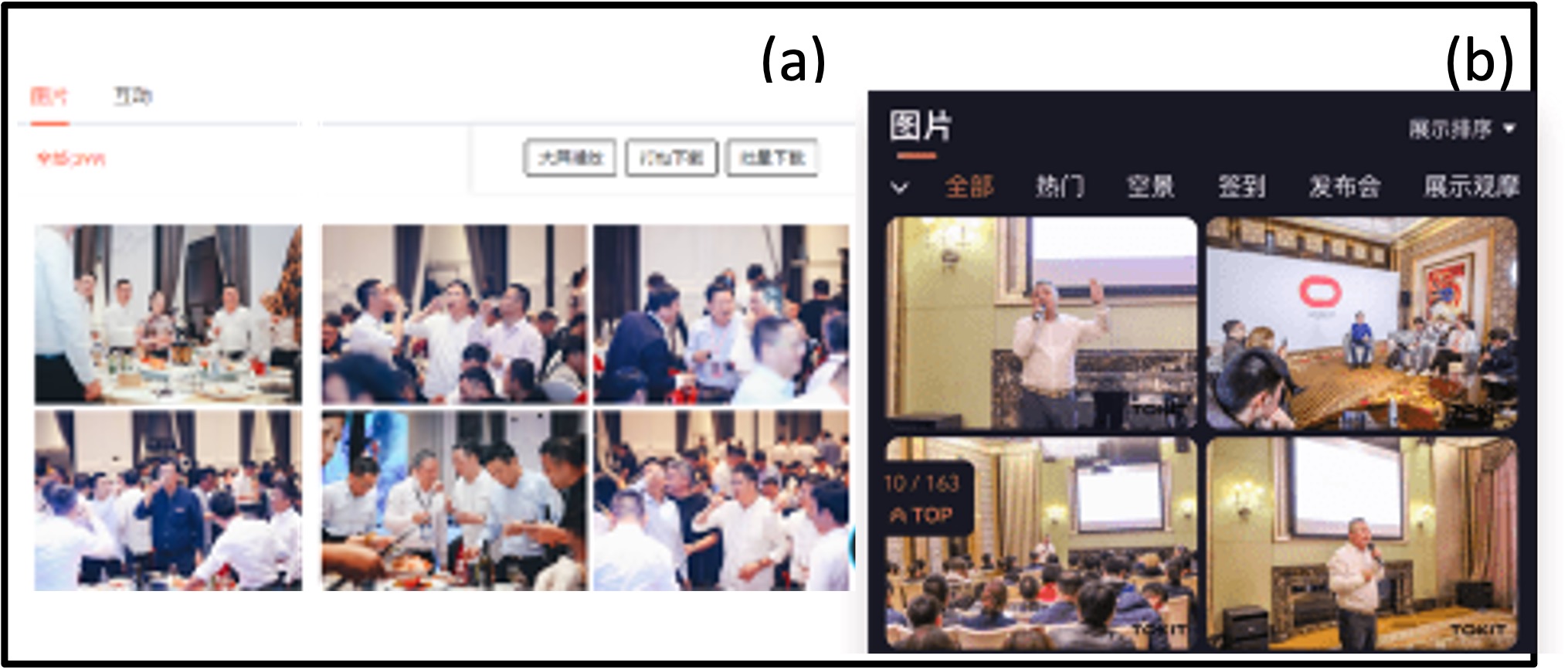} 
\caption{Images from 2 events on 2 photo live platforms: (a)photoplus, (b) xxpie. }
\label{fig:Example_ui}
\end{figure}

\begin{figure}
\centering 
\includegraphics[width=3.5in]{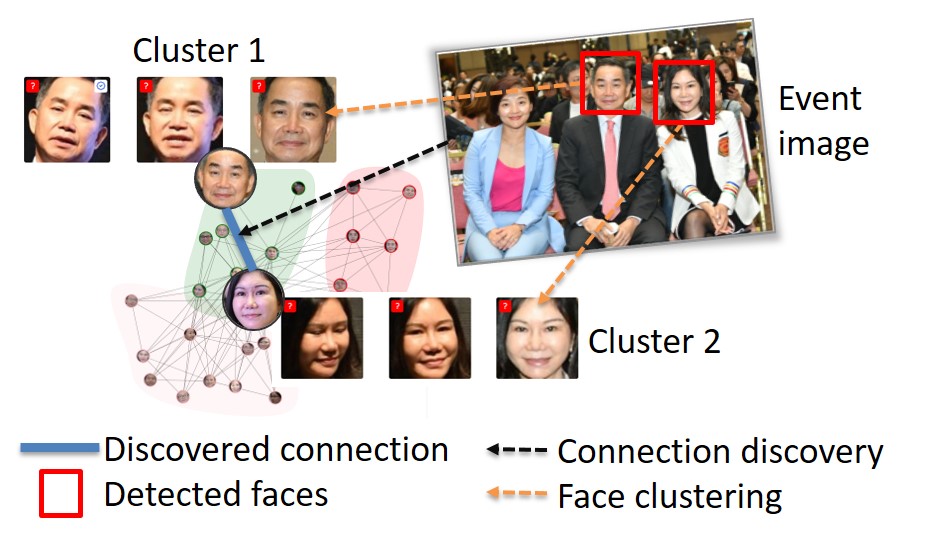} 
\caption{Social graph, clusters and images.}
\label{fig:socialGraph_cluster_image}
\end{figure}

\noindent
Social events are common and important for businesspeople to build connections and create new business opportunities in person \cite{zhao2015social, li2017social}. 
Event organizers usually hire photographers to take images which reveal information about the participants' connections. 
Unlike social media images, participants allow to be photographed at social events. 
On social media, users must confirm a connection among them, but in event images, participant connections are discovered without the needs of confirmation of the participants, and even if the participants do not have a social media account. 
Fig. \ref{fig:Example_ui} shows images from two photo live websites, photoplus\footnote{www.photoplus.cn} and xxpie\footnote{www.xxpie.com}. 
These websites provide a platform for event organizers, participants and photographers. 
Photographers can upload images to the platform in real-time, and participants and organizers can access the images immediately after they are taken.
Participants talk, shake hands and exchange name cards during an event. 
It has been shown that social graphs can be discovered those interactions on the event images, and the social graphs follow a power-law distribution, and participants form communities \cite{cheung2021discovering}, which is similar to those found on social media. 
Thus, social network analytics (SNA) \cite{marin2011social} can be conducted with the discovered connections.
As the faces have to be identified\cite{cheung2021discovering}, it is not clear if the framework can be applied if faces are not identified that a clustering process is required. 
\\
\indent
However, some unique issues must be addressed when extracting connections from event images. 
For example, it is possible to identify faces from a single image \cite{zhang2016joint} and the possible relationships between them \cite{xia2012understanding, li2017dual} on images such as a selfie, but it is unclear how to predict the relationship between two participants. 
This is because some participants are not known in advance, such as newcomers and walk-ins, so it is not possible to recognize them for identifying relationships. 
Fig. \ref{fig:socialGraph_cluster_image} shows the solution proposed in the paper for detecting connections from event images, and the relationship between clusters, faces, and connections in this paper. 
When faces are identified, they are grouped into different clusters and their relationships are inferred. 
As a result, the connections among participants can be discovered from the interactions on the images. 
Discovering connections from the event images requires face detection, face clustering, and interaction detection. 
However, existing face clustering algorithms are designed for general purposes, so they do not use information beyond an image, such as the social connections among the faces on the image. 
It is therefore desirable to utilize different information on event images and their participants to improve face clustering, such as the co-occurrence \cite{lan2012social}.
\\
\\
\indent
This paper mainly contributes the following: 
1)~proposed a framework for discovering connections from event images; 
2)~implemented a face clustering method for conducting face clustering using the framework; and 
3)~tested the effectiveness of the clustering method with real data from over 3,000 event participants, and demonstrating that the proposed method outperforms existing clustering methods.
The paper is organized as follows: Sec. \ref{sec:related_works} covers related works; 
Sec. \ref{sec:connection_discovery} introduces and formulates the system for discovering connections; 
Sec. \ref{sec:analytics} analyzes the discovered connections from faces, and summarizes the limitations of existing face clustering algorithms; 
Sec. \ref{sec:algorithm} discusses the proposed face clustering methods from event images based on the analytic, followed by the experimental results in Sec. \ref{sec:results}; and Sec. \ref{sec:conclusion} concludes the paper.
\\
\section{Related Works}
\label{sec:related_works}
\noindent
There are numerous applications related to social graphs, ranging from recommendation \cite{konstas2009social} and tie prediction \cite{wang2011human} to group decision making \cite{zhou2019social}. 
Social graphs are particularly useful for online platforms like Facebook, where users are motivated to provide easily accessible and exploitable social graphs to the operators. 
However, the social graphs are not always accessible, those applications make the graphs inaccessible due to the privacy concerns, or exclusive for the operators of the online platforms.
One approach to discovering connections is to use user-shared images \cite{cheung2019detecting}. 
Relationships such as husband-wife, parents-children, siblings, and grandparents-grandchildren can be discovered from image sets \cite{sun2017domain, wang2010seeing, singla2008discovery, dai2015family}. 
Although most works focus on telling relationships in a single image, it is not clear how to detect a relationship from a set of images. 
As an event will produce a set of images, it is necessary to recognize faces in images, and relationships are discovered based on the images that two participants have co-occurrence \cite{cheung2021discovering}.
A lot of research has been done on face recognition using deep learning \cite{wang2018deep}. 
Faces are first detected from images, and multi-task cascaded convolutional networks \cite{zhang2016joint} can be used for detection. 
Faces are then represented as a vector \cite{ding2016comprehensive, schroff2015facenet}. 
To enhance recognition, faces can be enhanced through methods such as face frontalization \cite{zhu2016face, huang2017beyond, jourabloo2017pose} and face alignment \cite{sagonas2015face, guo2018stacked}. 
Relationships in images help with recognition \cite{oh2016faceless}. 
However, many participants do not provide their identities, making direct recognition impossible.
\\
\indent
Therefore, to discover relationships from images with unknown faces, the first step is to use face representation or embedding, and then tag the faces of the same person with a unique label. 
This is a clustering process \cite{yen2009cluster, dubes1988algorithms, shi2018face, liu2021learn}, in which faces are grouped based on their embeddings. 
However, as the faces of different persons could be similar due to lighting conditions and the face angle, it is desirable to utilize the information provided by the images \cite{otto2017clustering, wang2019linkage, papagiannopoulou2014concept}. 
Even unrecognizable faces can be grouped according to their identity, and relationships can be obtained. 
The information of those unrecognized faces can be input by the event organizer later. 
Thus, the performance of clustering is critical for the discovery but using generic clustering algorithms alone cannot provide a good result \cite{wu2010improving}. 
Algorithms like $K$-means require the number of participants to be known in advance, which is not possible in events as participants may be walk-ins. 
Density-based methods \cite{rodriguez2014clustering, guo2020density} do not solve the problem as the faces of different participants may have high similarities in terms of embedding due to lighting conditions and the angle of the faces to the camera.
\\
\indent
Different image information, including time \cite{zhao2006automatic}, human attributes \cite{shen2017learning, lu2018attribute, zhang2014context}, and clothing \cite{zhang2014context} are used for clustering. 
This information is used to decide whether a face belongs to a cluster or detect whether a face in a cluster does not belong to it. 
With the fast-growing technologies of deep learning-based face recognition, face attributes such as ages can be effectively captured \cite{schroff2015facenet}. 
Different from the input in \cite{zhang2014context}, images are captured by the event photographers, and faces of the same participant could be frontal/non-frontal. 
However, these methods mainly focus on images from social media, and some information is not applicable to event images or not available to images from social media.
For event images, they are taken in a short period, and one can assume that the face and cloth remain consistent during the event. 
To cluster faces from event images, co-occurrence is used \cite{lan2012social, zhang2014context} to detect whether the same participant appears on the same image twice \cite{xia2014face}. 
As photographers take images of the same participants in a short period from different angles, faces on images taken in a short period are more likely to be from the same participant. 
Hence, a framework is proposed to utilize such knowledge related to images and events for a better performance on clustering faces from event images than conventional approaches.
This paper extends \cite{cheung2021discovering} in the following ways:
1)~collected and tagged a dataset of 40000 faces with over 3000 participants; and,
2)~conducted analytics on the event images to showcase how existing face clustering algorithms do not perform well;
3)~proposed a framework for clustering faces, and tested it on real data to show effectiveness.
\\
\indent
In the coming sections, the framework of connection discovery first introduced to tell that a good face clustering is needed for connection discovery. 
Hence, The next section conducts analytics on why the existing methods fails, and the reason for designing steps for existing face clustering algorithms on event images. 
Sec. \ref{sec:algorithm} is the proposed clustering steps based on the analytics 

\begin{figure}
\centering 
\includegraphics[width=3.5in]{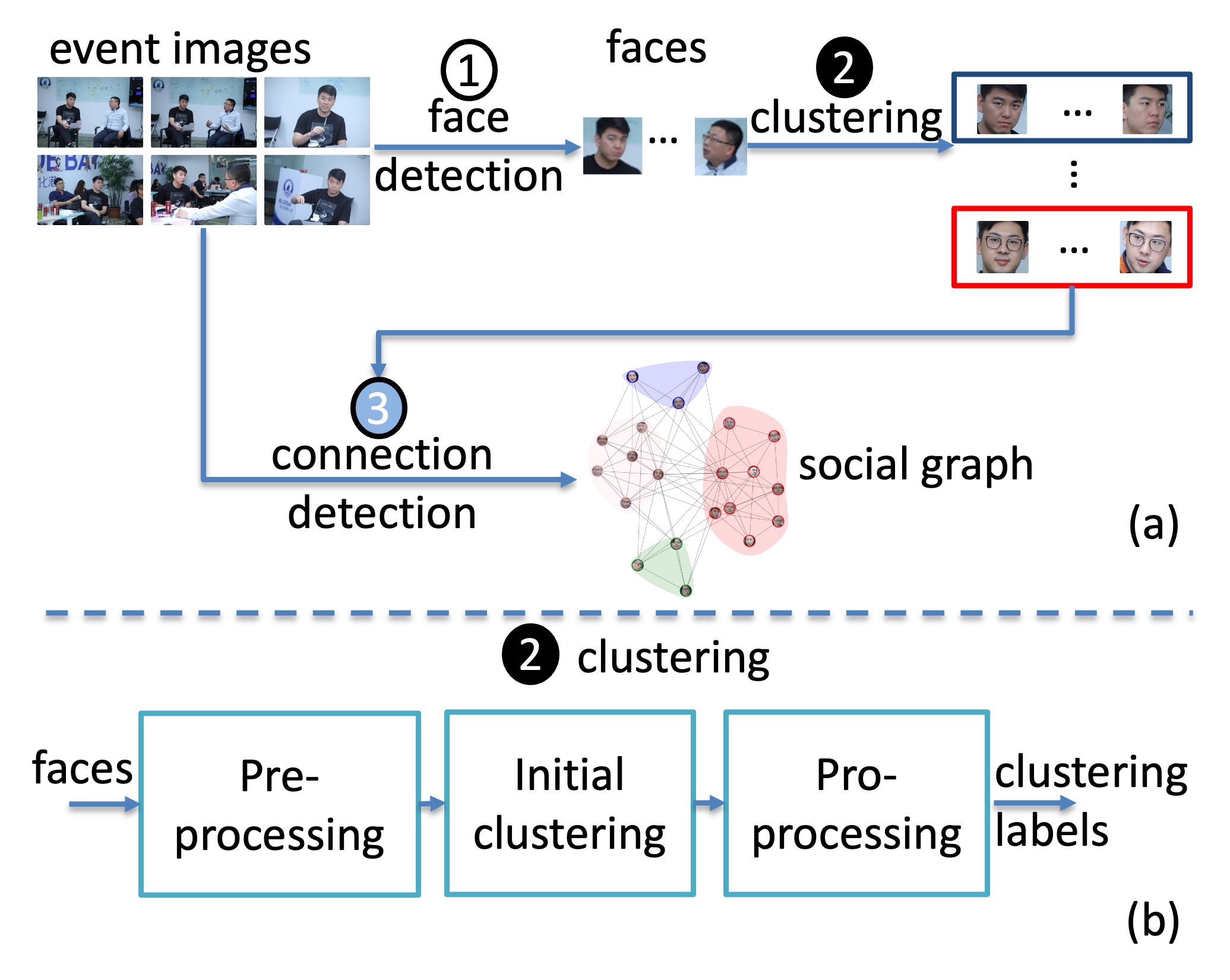} \caption{The proposed system: (a) system flow. (b) clustering process}
\label{fig:systemFlow}
\end{figure}
\section{A Framework of Connections Discovery}
\label{sec:connection_discovery} 
\noindent 
This section introduces the framework for discovering connections between participants in an event. 
Figure \ref{fig:systemFlow} (a) shows the flow of the framework, which begins by detecting faces from images and clustering them accordingly. 
The clustering results are then used to discover connections.

\subsection{Face Detection and encoding}
\noindent 
The first step is to detect faces in the images and encode them, as shown in step 1 of Figure \ref{fig:systemFlow}. 
The faces are encoded in a way that faces belonging to the same person are likely to have similar embeddings, and vice versa.
Different face detection algorithms can be applied, such as multitask cascaded convolutional networks \cite{zhang2016joint} and FaceNet \cite{schroff2015facenet} encoding.

\subsection{Face Clustering}
\noindent 
Face clustering groups faces based on their embeddings, such that faces of the same participant are in the same cluster, and the faces of different participants are in different clusters, as shown in Figure \ref{fig:systemFlow} (b). 
Clustering algorithms such as $K$-means or DBSCAN \cite{rodriguez2014clustering} can be applied to obtain the labels of faces.

\begin{figure*}
\centering 
\includegraphics[width=5.5in]{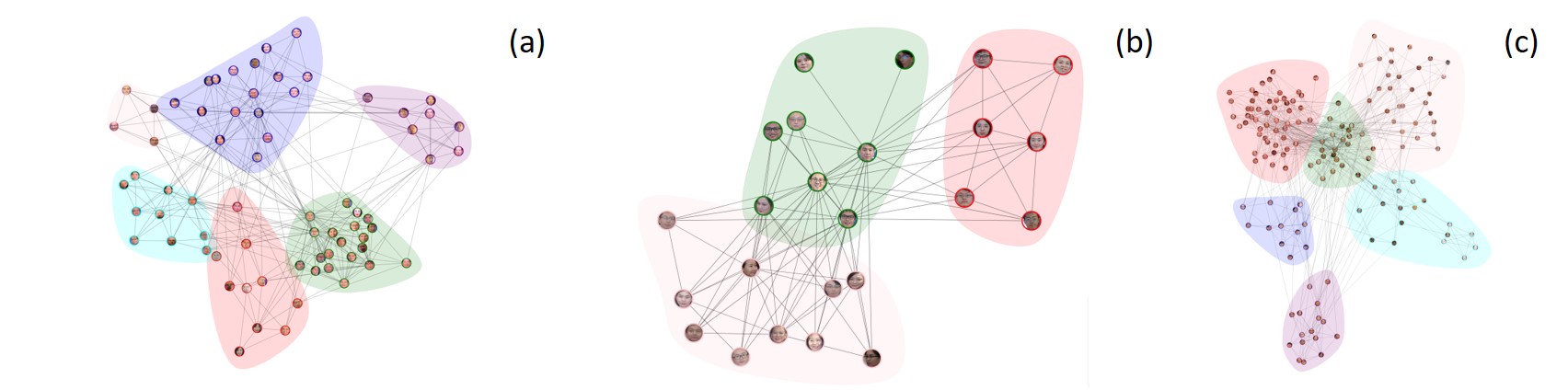} \caption{Social Graphs built from different events. in (a) an art gallery, (b) a product launch, (c) a format dinner.}
\label{fig:socialGraph}
\end{figure*}
\subsection{Clustering for Connection Detection}
\label{Discovering Connections} 
\noindent 
The connections are discovered based on the co-occurrence of two persons, as shown in step 3 of Figure \ref{fig:systemFlow}. 
As shown Fig. \ref{fig:socialGraph_cluster_image}, each cluster represents a participant, and if two faces from different clusters are presented in the same image and the participants are interacting, a connection is formed between the two clusters. 
When faces from all images of an event are analyzed, the connections among participants are discovered and the social graph is formed.
\\
\indent
As shown in \cite{cheung2021discovering}, the social graphs formed with only co-occurrence follow the properties of social media graphs well, but they require manual labeling for the identity of faces. 
To discover connections among participants, it can be formulated as $P(C_{i,j}=1|I_0,I_1,...I_N)$, the probability of two participants, $i$ and $j$, being connected given $N$ images in an event. 
Let $\mathbf{I}(i,j)$ be the set of images that the two participants have interacted during the event, the probability becomes:
\begin{equation}
\label{eq:connection}
P(C_{i,j}=1|I_0,I_1,...I_N) = 
\begin{cases}
 & 1 \text{ if }|\mathbf{I}(i,j))|>0\\ 
 & 0 \text{ if } otherwise
\end{cases}
\end{equation}
Thus, it is desirable to design a clustering method for connection discovery.
In the next section, analytics will be conducted to understand how to perform clustering for connection discovery.

\section{Analytics of Event images}
\label{sec:analytics} 
\noindent
This section presents analytics on event images and connection discovery. 
The first part showcases cases wherein faces are correctly labelled with an identity, and the corresponding social graphs from the discovered connections. 
The second part highlights the difficulties of face clustering, and the need for a better face clustering algorithm.

\subsection{Social Graph}
\noindent
To demonstrate the effectiveness of connection discovery using event images, an experiment is conducted using the connection discovery framework.
Instead of using a clustering algorithm, the detected faces are manually clustered, and hence, are all properly labelled. 
The connections are discovered accordingly, and the social graphs formed with discovered connections are shown in Fig. \ref{fig:socialGraph}. 
Figs. \ref{fig:socialGraph} (a-c) correspond to events in an art gallery, a product launch, and a formal dinner with different numbers of participants. 
It is observed that they have similar properties to a social graph from online social media. 
The communities are highlighted in different colours, and some persons are more important than others while most of them are not important. 
People are clustered into different communities, and they follow some important patterns as the social graphs from online social media. 
However, to discover the connection, it is necessary to have an accurate face clustering result. This section discusses cases that the conventional clustering algorithm fails to handle.
\begin{figure*}
\centering 
\includegraphics[width=5.5in]{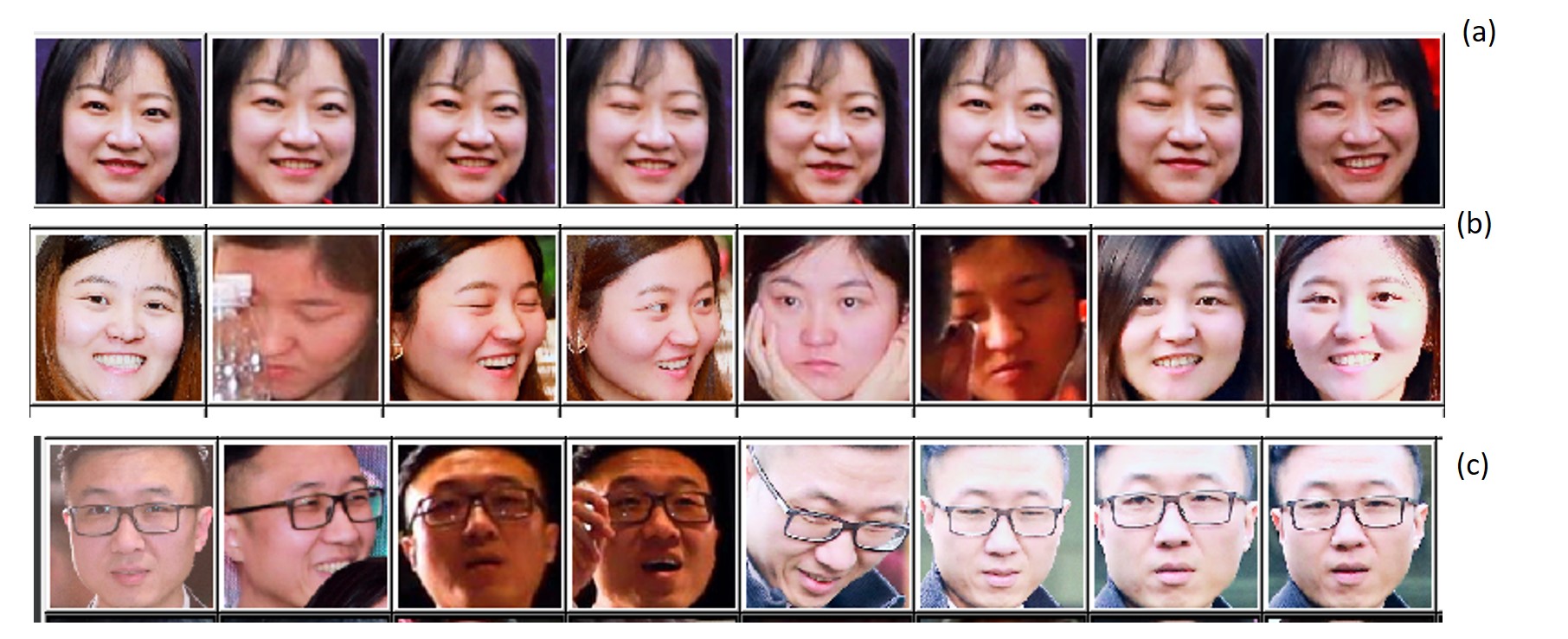} \caption{Example results of clustered faces of 3 different participants of the same event: (a) not much changes, (b) with different angles and emotions, (c) from indoor and outdoor.}
\label{fig:clusteringExample}
\end{figure*}
\begin{figure}
\centering 
\includegraphics[width=3.5in]{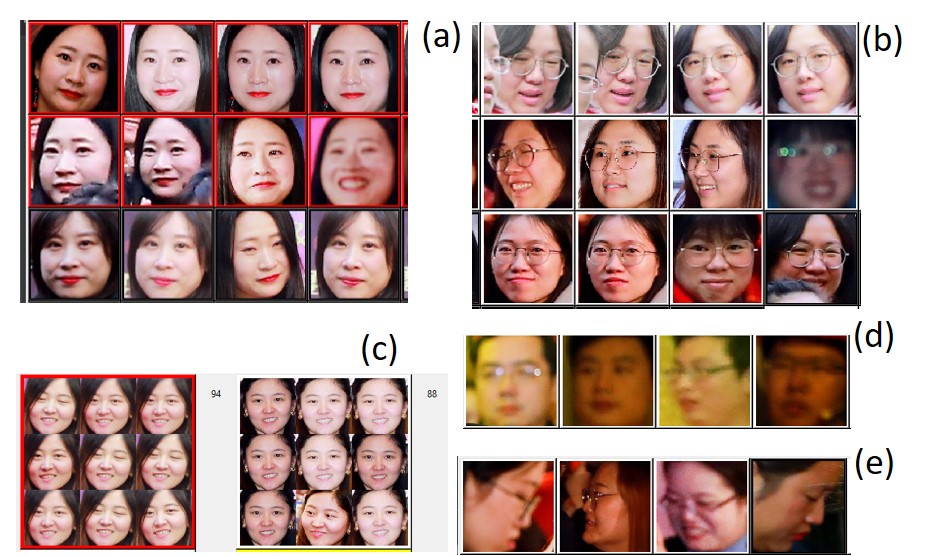} \caption{Limitations of DBSCAN: (a) \& (b) contain faces of more than 1 persons; (c) 2 clusters with high similarity; (d) \& (e) blurred and side faces from different participants are grouped.}
\label{fig:Failed}
\end{figure}
\subsection{Face Clustering}
\noindent
The relationship between clusters and faces is essential for connection discovery, as shown in Fig. \ref{fig:socialGraph_cluster_image}. 
Without a good clustering result, many wrong connections will be discovered, while many will be missed. 
To understand the difficulties of clustering event images, Fig. \ref{fig:clusteringExample} shows examples of clusters from an event. 
In Fig. \ref{fig:clusteringExample} (a), the faces on the images are taken during a short period, and they look very similar. 
In Fig. \ref{fig:clusteringExample} (b), the faces are taken at different times and could have different brightness, angles, and emotions. 
In Fig. \ref{fig:clusteringExample} (c), it is a combination of the two, with some images taken indoors and some taken outdoors.
As faces are taken during an event, their ages and some face attributes, such as glasses, are consistent.
\\
\indent
Analytics is conducted to showcase the limitations of using a conventional face clustering algorithm, and the results are shown in Fig. \ref{fig:Failed}.
In this work, the faces are encoded using pre-trained models that generate 128-dimensional embedding of FaceNet, and then they are clustered by DBSCAN\cite{rodriguez2014clustering}, and selected to showcase the limitations of the current algorithm. 
FaceNet is a deep learning-based face recognition by Google that encodes a detected face into a face embedding.
Figs. \ref{fig:Failed}(a) and (b) show clusters that contain faces of more than one person. 
Although the faces are obviously from two participants and are not blurred, they have high similarity in terms of both embeddings. 
Fig. \ref{fig:Failed}(c) shows another type problem. 
The faces of the same participants are put into different clusters, as they have slightly different appearances and the heir styles.
Even they have very high similarity, the faces are in 2 clusters.
Figs. \ref{fig:Failed}(d) and (e) show other types of problems. 
Faces that belong to different participants are grouped into a cluster. 
The reason behind this is that they are too blurred for the embedding, and they are grouped into the same cluster. 
Based on these observations, a clustering method is proposed to improve the face clustering that utilizes event information for the pre-processing and pro-processing of any clustering algorithm. 
The method is designed based on the inputs available from event images.

\section{Proposed Face Clustering Method}
\label{sec:algorithm} 
\noindent 
This section discusses the proposed clustering method.
Pre-processing operations are installed to enhance the clustering of event images.
The faces are then clustered using a conventional clustering approach, followed by operations using domain specified features to improve the performance of the clustering.
\subsection{Pre-processing}
\subsubsection{Face Filtering}
\noindent
The aim of face filtering is to filter out unrecognizable faces, such as blurred faces. 
A classifier is trained to distinguish between suitable and unsuitable faces for clustering.
Faces that are not unrecognizable are labelled as '0', and the rest is labelled as '1'.
The classifier is trained using face embedding and faces below a threshold are rejected. 
Note that a blurred/dark/bright face can be labelled as '1', if it can be recognized by humans.
This is necessary because when participants are talking, images from different angles can be taken.
It is a necessity as when participants are talking, the images taken may be from different angles, and these images should be kept.

\subsubsection{Time Grouping (time)}
\noindent
As event images often have similar settings, time grouping labels two faces as the same person if the difference between the time taken is small.
For example, while 2 participants are talking, the photographers would take images around them.
Although the appearance of the images may be very different, it is likely that they are images of the same set of participants but taken from different angles.
Hence, this operation is to label 2 faces as the same person using a lower threshold, if the difference in the time taken is small.
\subsubsection{Check Duplication (check)}
\noindent
Unlike social media, people share only the best images even if many are taken in one second (see Fig. \ref{fig:clusteringExample} (a)). 
Raw data may therefore contain many duplicate images, affecting the clustering process \cite{costa2010incremental}. 
Duplicates are detected before clustering as pre-processing.
If taken within a few seconds, images with similar faces are considered duplicates and only one is used for clustering. 
The faces on duplicated images are recognized after the clustering results are ready.

\subsection{Initial Clustering}
\noindent
Density-based algorithm, DBSCAN\cite{rodriguez2014clustering}, is used to cluster the faces. 
DBSCAN views clusters as areas of high density separated by areas of low density. 
Clusters can therefore be any shape, unlike $k$-means which assumes convex shapes. 
The operations work with any clustering methods and better methods improve the results. 
Post-processing follows to further enhance the results.

\subsection{Pr-processing}
\subsubsection{Co-occurrences (same)}
\noindent
In event images, the same participant cannot appear on the same image \cite{zhang2014context}. 
Unlike shared images from social media, event images are taken by photographers, and multiple images are not merged into a single image before sharing.
This assumption is valid as images are taken by photographers and not merged before sharing. 
The operation is conducted using hierarchical clustering when 2 faces from the same image are grouped into the same cluster, in which the 2 faces cannot be put into the same cluster\cite{xia2014face}.
This ensures two participants even with similar faces are split into two clusters.

\subsubsection{$K$-nearest Neighbour (knn)}
\noindent
Although most faces can be recognized by the similarity of their embeddings, there are some cases where faces are similar but fall below the recognition threshold. 
To recognize more faces, a $K$-nearest Neighbour ($K$NN) algorithm is used. 
For a face that is not in any cluster, the top 5 faces with the highest similarities are located, and a majority vote is conducted to decide which cluster to assign the face to. 
If at least 4 out of the 5 faces have the same label, the face is labeled with that same label. 
This operation allows a face to be assigned to a cluster even if there is a relatively small similarity to faces in another cluster.

\subsubsection{Cluster Checking (neghigh)}
\noindent 
After clustering with the proposed framework, there may still be clusters with faces belonging to different people. 
These faces are of low quality but not low enough to be rejected. 
For example, they could be clusters containing side faces, as shown in Fig. \ref{fig:clusteringExample} (e). 
The faces in the cluster have high similarities as a result. 
This operation assumes that for a participant, there exist some faces that have a good face score, i.e., faces that are taken clearly. 
This is a valid assumption as most participants will capture and take group photos with some high face score faces. 
The goal of this operation is to remove those clusters. 
It is conducted by locating the face with the highest face score in each cluster and rejecting those clusters with no face with a high face score. 
Once the pre-processing is completed, the results are ready for connection discovery.

\begin{figure*}
\centering 
\includegraphics[width=5.5in]{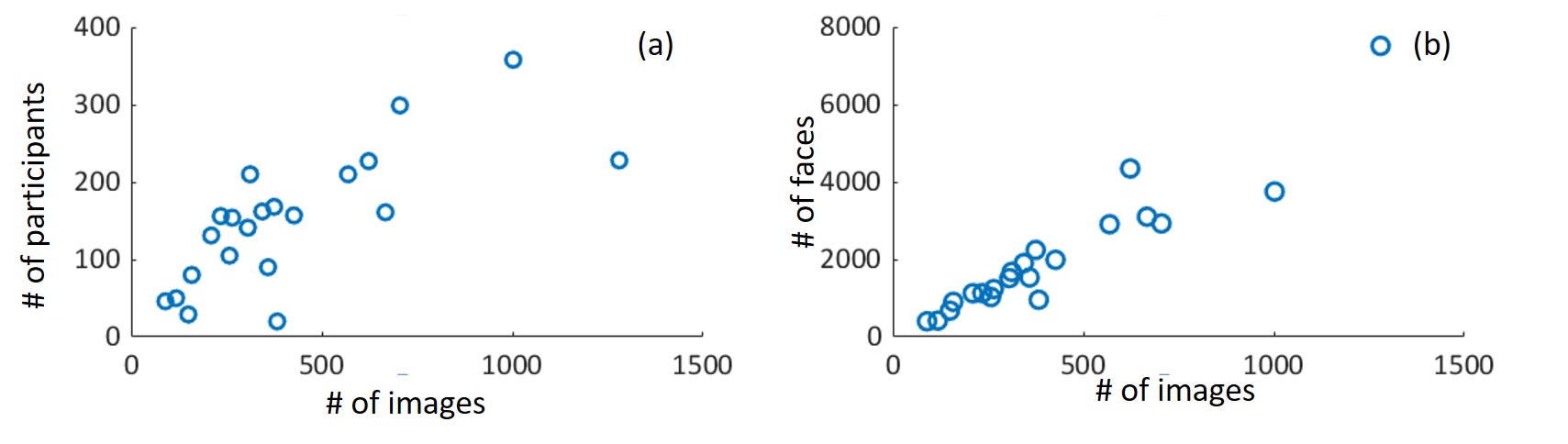} \caption{Statistic of the events involved: (a) \# of images vs. \# of participants, (b) \# of images vs. \# of faces.}
\label{fig:socialGraphStatistic}
\end{figure*}

\section{Experimental Results}
\noindent 
\label{sec:results} 
This section discusses the preparation of data, as well as the details of the settings and results of the experiments. 
The proposed algorithm is compared with other clustering algorithms, followed by the results on the effectiveness of different operations. 
The robustness and how different encoding methods affect the results are also discussed.

\subsection{Experimental Settings}
\noindent 
Data was collected from pailixiang.com, a Chinese photo live platform. 
The event organizer uploads event images to the website during the event, and they are shared publicly online.
Images do not come with information other than the upload time and the number of views. 
As there is no identity information available, faces are labeled with the identity manually using a custom-developed software. 
After manual labeling, there are over 3,000 participants labeled from over 40,000 faces and 8,837 images in the data set.
The details of the tagging process and the software can be found in the appendix. 
The relationships between the number of images, participants and faces are shown in Fig: \ref{fig:socialGraphStatistic}.
It is observed that they have a linear relationship that an event with a larger number of images has more participants and faces.
\\
\indent
In the experiment, the faces are encoded using pre-trained models that generate 128-dimensional embedding of FaceNet, a deep learning-based face recognition by Google that encode a detected face into a face embedding.
The embedded faces are then clustered with the proposed methods, and the clustering results are compared with the ground truth, the labels on the faces based on the identity.
There are 5 baseline methods implemented, in which the same embedded faces are used as the inputs but using a different clustering algorithms.
The first one is DBSCAN \cite{rodriguez2014clustering, lin2018deep}, with eps = 50, min samples to be 3 and metric to be euclidean distance.
The second one is $K$-means and the third one is spectral clustering, in which the number of clusters is set to 50.
The forth one is random, in which the faces are randomly assigned to 50 clusters.
The last one is the approach designed for clustering hundreds of millions of faces using Rank-Order clustering \cite{otto2017clustering}. 
The same settings as in \cite{otto2017clustering} are applied \footnote{available:github.com/varun-suresh/Clustering}, with the number of nearest neighbors equal to 10. 
The reason is that when the number of nearest neighbors is high, most of the faces belong to the same cluster.

\subsection{Performance Measurement}
\noindent 
An important measurement is whether the faces are correctly clustered.
As there is no direct mapping between the clusters and participants, it is not possible to compute the precision $p$ and recall $r$ for a single face, as in face recognition. 
Hence, this paper considers face clustering as face pair binary classification problem in measurement, in which if a face pair is predicted as class 1 if they are clustered into the same cluster, and is class 0 is they are not in the same cluster.
If 2 faces are labelled to be the same person, they are as  class 1, and vice verse.
With these 3 values, the precision $p$ and recall $r$ can be calculated by summing up all possible face pairs:
\begin{equation}
\label{eq:precision}
p = \sum_{f_a \neq f_b} \dfrac{{T^\prime}_p(f_a,f_b\in\ {\bar{\mathbf{F}}}_{i\prime})}{{T^\prime}_p(f_a,f_b\in\ {\bar{\mathbf{F}}}_{i\prime})+{F^\prime}_p(f_a,f_b\in\ {\bar{\mathbf{F}}}_{i\prime})} \\
\end{equation}
\begin{equation}
\label{eq:recall}
r =   \sum_{f_a \neq f_b} \dfrac{{T^\prime}_p(f_a,f_b\in\ {\bar{\mathbf{F}}}_{i\prime})}{{T^\prime}_p(f_a,f_b\in\ {\bar{\mathbf{F}}}_{i\prime})+{F^\prime}_N(\ f_a\in\ {\bar{\mathbf{F}}}_{i\prime},\ f_b\in\ {\bar{\mathbf{F}}}_{j\prime})} \\
\end{equation}
Where ${T^\prime}_p(f_a,f_b\in\ {\bar{\mathbf{U}}}_i\prime)$, ${F^\prime}_p(f_a,f_b\in\ {\bar{\mathbf{U}}}_{i\prime})$ and ${F^\prime}_N(\ f_a\in\ {\bar{\mathbf{F}}}_{i\prime},\ f_b\in\ {\bar{\mathbf{F}}}_{j\prime})$ are the true positive, false positive and false negative, respectively.
The details of them can be found in the appendix.
\\
\indent
As the main goal of the system is to discover the connection and the social graph among the participants, hence another measurement is to measure whether an algorithm can discover the top 10 participants with the highest number of connections.
To count the number of clusters that can be mapped to participants in the ground truth, Eq. \ref{eq:count_rs} is defined as:
\begin{equation}
\label{eq:count_rs}
a( \mathbf{F}_{i}, \mathbf{\bar{F}}_{i'})  = 
\begin{cases}
 & 1 \text{ if } J(\mathbf{F}_{i}, \mathbf{\bar{F}}_{i'})>t_J\\ 
 & 0 \text{ if } otherwise
\end{cases}
\end{equation}
where $J(.,.)$ is the Jacob similarity between the set of faces of a participant $i$, $\mathbf{F}_{i}$, and the set of faces of a cluster, $\mathbf{\bar{F}}_{i'}$.
As a cluster may mix faces of different participants, or miss some faces of that participant, $t_J$ is the threshold to decide whether the two sets are mapped.
In the paper, the faces in a cluster has to be with at least 0.8 in terms of Jacob similarity in order to be counted as that participant.
Hence, $rs$, is defined as the percentage of the top 10 participants discovered into a cluster for each event:
\begin{equation}
\label{eq:rs}
rs(U_{top}) = \sum_{i \subset U_{top}, i' \subset \overline{U}} \frac{a( \mathbf{F}_{i}, \mathbf{\bar{F}}_{i'}) }{|U_{top}|}
\end{equation}
where $U_{top}$ is the top 10 participates in the ground truth, and $\overline{U}$ is the set of clusters after clustering.
A better algorithm gives a higher $rs$.
\begin{figure}
\centering 
\includegraphics[width=3.5in]{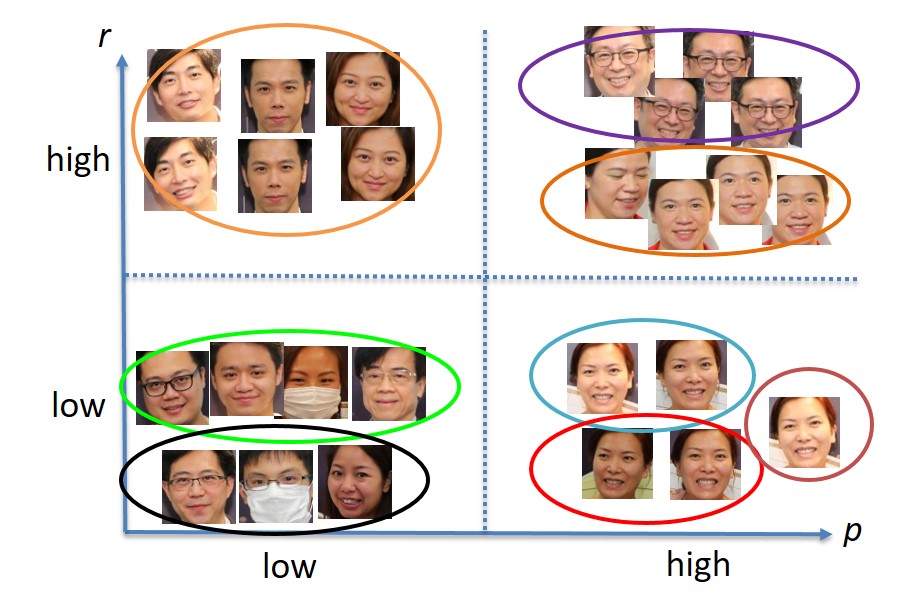} \caption{different results for precision and recall}
\label{fig:highlowscore}
\end{figure}
\subsection{Comparison with Baselines}
\begin{figure*}
\centering 
\includegraphics[width=5.5in]{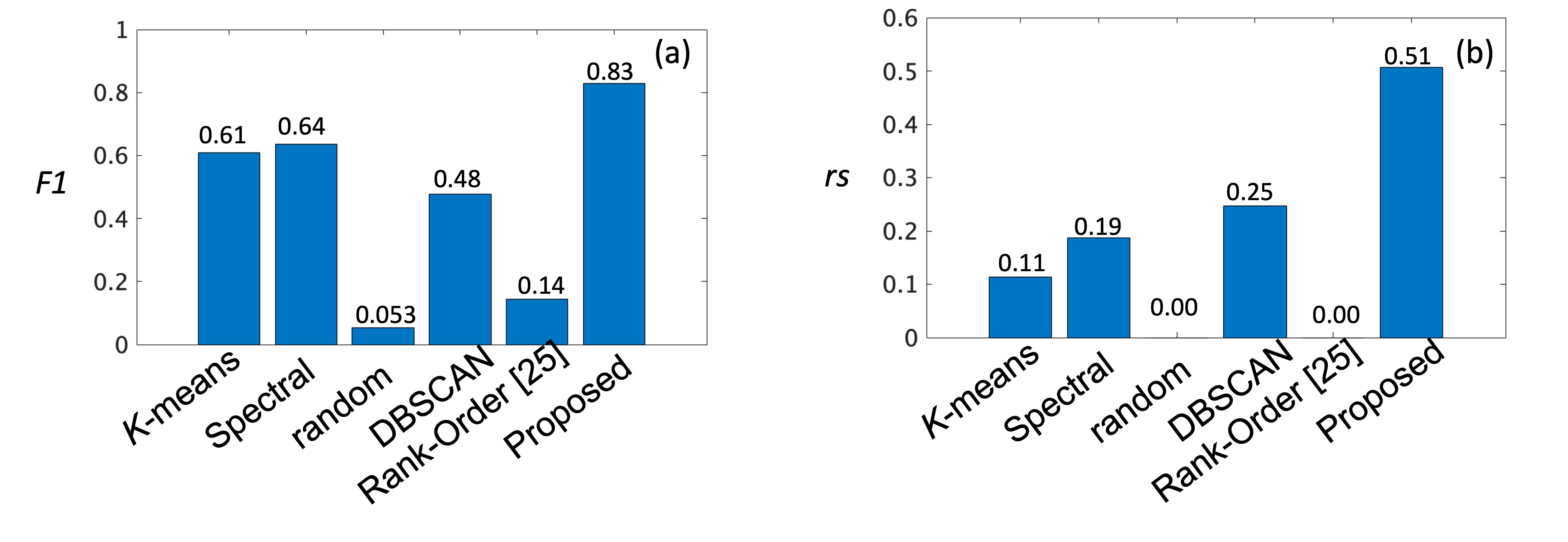} \caption{Results of the proposed algorithm: (a) $F1$, (b) $rs$.}
\label{fig:Result1}
\end{figure*}

\noindent
This section compares the proposed algorithm with other algorithms for face clustering and the result is shown in Fig. \ref{fig:Result1}. 
It is observed that the proposed algorithm outperforms other approaches in terms of $F1$ and $rs$.
This means that the proposed algorithm is better at clustering faces and correctly identifying important participants. 
The $F1$ score is around 0.8. 
The results also highlight the importance of using $rs$. 
While $K$-means and spectral clustering perform better than DBSCAN in terms of $F1$ score, they may also include faces of unimportant participants. 
The proposed approach outperforms Rank-Order clustering as well. 
One reason for this is that Rank-Order clustering is designed for millions of images, whereas there are fewer images in an event.
\subsection{Effects of different pro-processing methods}
\begin{figure*}
\centering 
\includegraphics[width=5.5in]{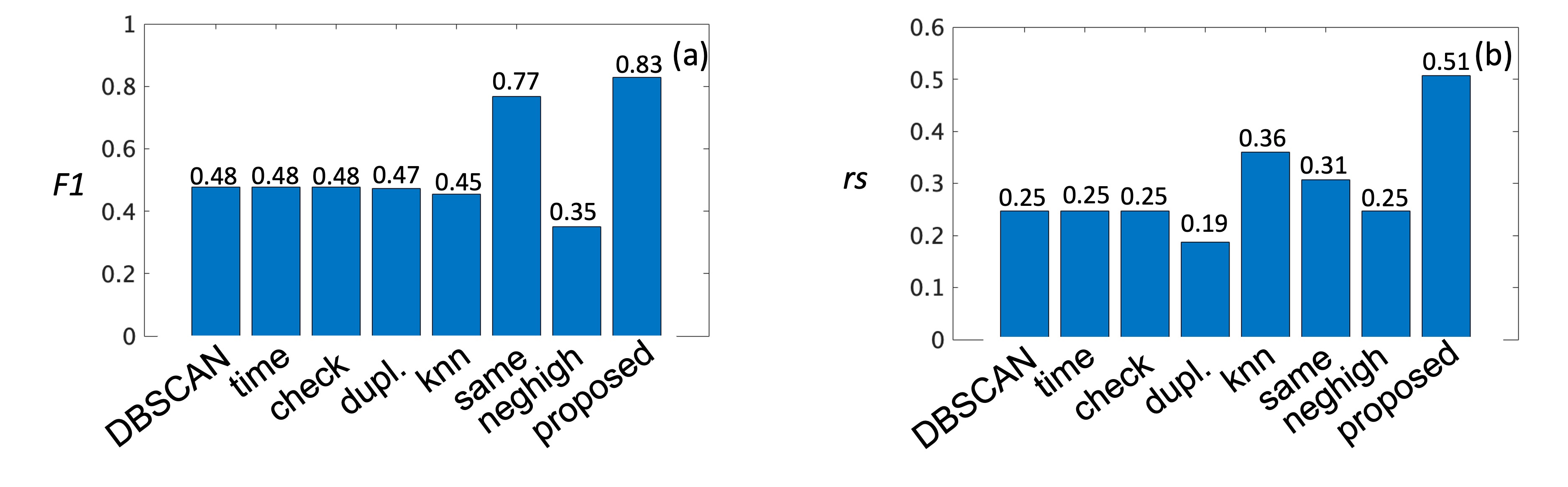} \caption{Results of different pro-processing, applied individually with initial clustering of DBSCAN: (a) $F1$, (b) $rs$.}
\label{fig:Result_prop_single}
\end{figure*}

\begin{figure*}
\centering 
\includegraphics[width=5.5in]{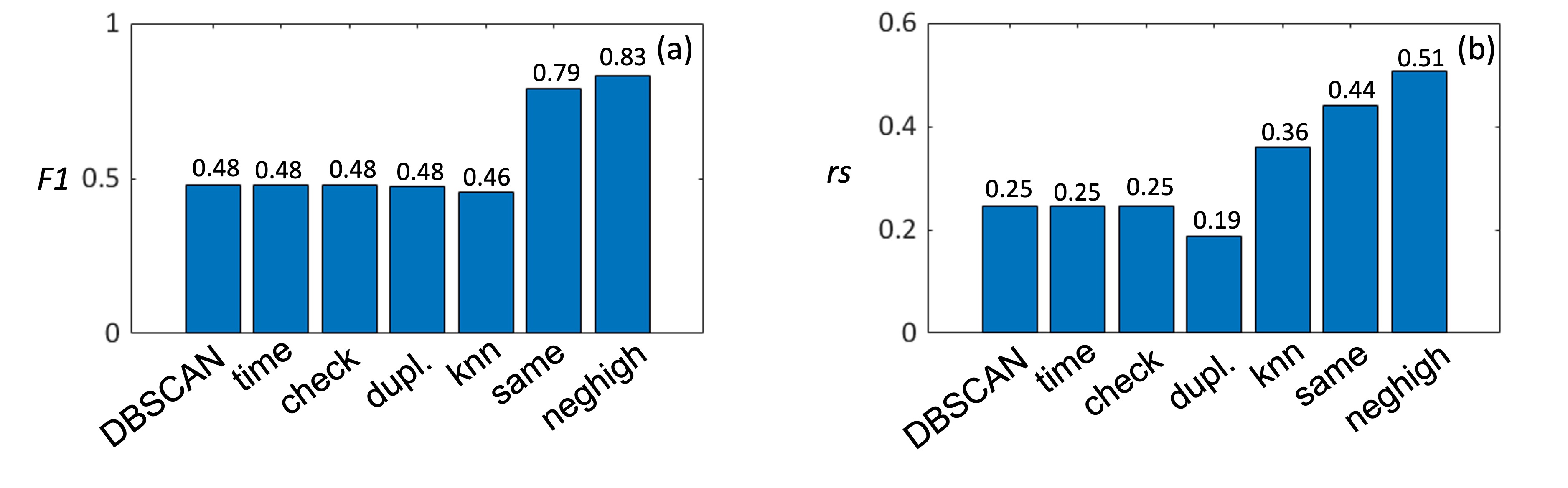} \caption{Results of different pro-processing, applied cumulatively with initial clustering of DBSCAN: (a) $F1$, (b) $rs$.}
\label{fig:Result_prop_cum}
\end{figure*}

\noindent
This section investigates the effects of different pro-processing, by only applying one effect at a time. 
Fig. \ref{fig:Result_prop_single} and Fig. \ref{fig:Result_prop_cum} show the result for applying individual and cumulative operations, respectively. 
It is observed that co-occurrence gives the most improvements in the result. 
By using knn, although the $F1$ drops as there are more wrongly clustered faces, $rs$ increases as more faces can be clustered correctly to pass the threshold for $rs$.
Fig. \ref{fig:Result_prop_cum} displays the accumulated results of the operations. 
It is important to note that the effects of different methods do not accumulate linearly, and their combinations are obtained through trial and error.
One of the reasons for this is that knn increases the number of faces in a cluster, making it more likely for them to surpass the threshold of $rs$, that is, when at least 0.8 in terms of Jacob similarity between the faces of a cluster and a participants. 
Therefore, while knn may wrongly recognize some faces into clusters, it contributes to correctly labeling the faces of important participants.
Furthermore, some operations such as neghigh do not improve $F1$ and $rs$ when applied individually, but show an improvement if applied cumulatively.
\subsection{Robustness}
\begin{figure*}
\centering 
\includegraphics[width=5.5in]{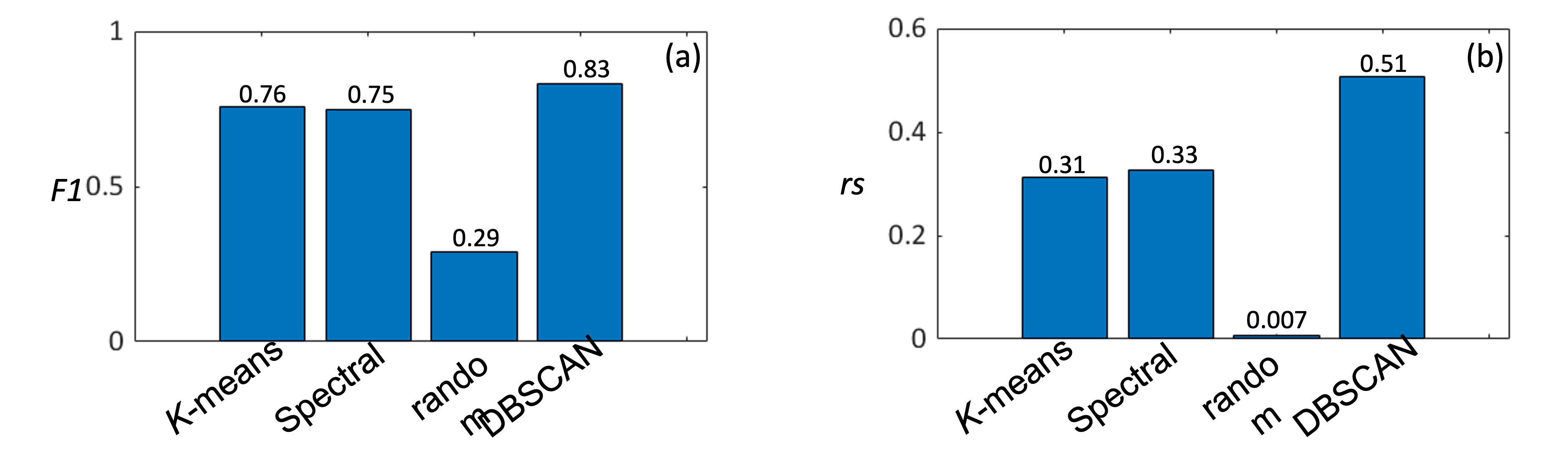} \caption{Results of different initial clustering algorithms: (a) $F1$, (b) $rs$.}
\label{fig:Robustness}
\end{figure*}

\noindent
This section investigates the robustness of the proposed algorithm by using different clustering algorithms in the initial clustering, followed by the same operations. 
There are four different algorithms, DBSCAN, $K$-means, spectral, and random approaches, as the initial clustering method before conducting different operations. 
The random approach is to give each face a random label after preprocessing, followed by the proprocessing. 
The results are shown in Fig. \ref{fig:Robustness}. 
The system is observed to be robust with different clustering algorithms, while DBSCAN gives the best result. 
In terms of $F1$, the three algorithms: DBSCAN, $K$-means, and spectral give similar performance, while DBSCAN is the best for $rs$. 
For Rand, it has the highest improvement after using the operations: 400\% improvement in terms of $F1$. 
For DBSCAN, an improvement of 73\% is observed. The results have proven that the proposed framework works well with different initial clustering algorithms, while a better algorithm will give a better result. 
Hence, any better clustering algorithms can be used as the initial ones, and a better result can be obtained using the proposed framework.
\begin{figure*}
\centering 
\includegraphics[width=5.5in]{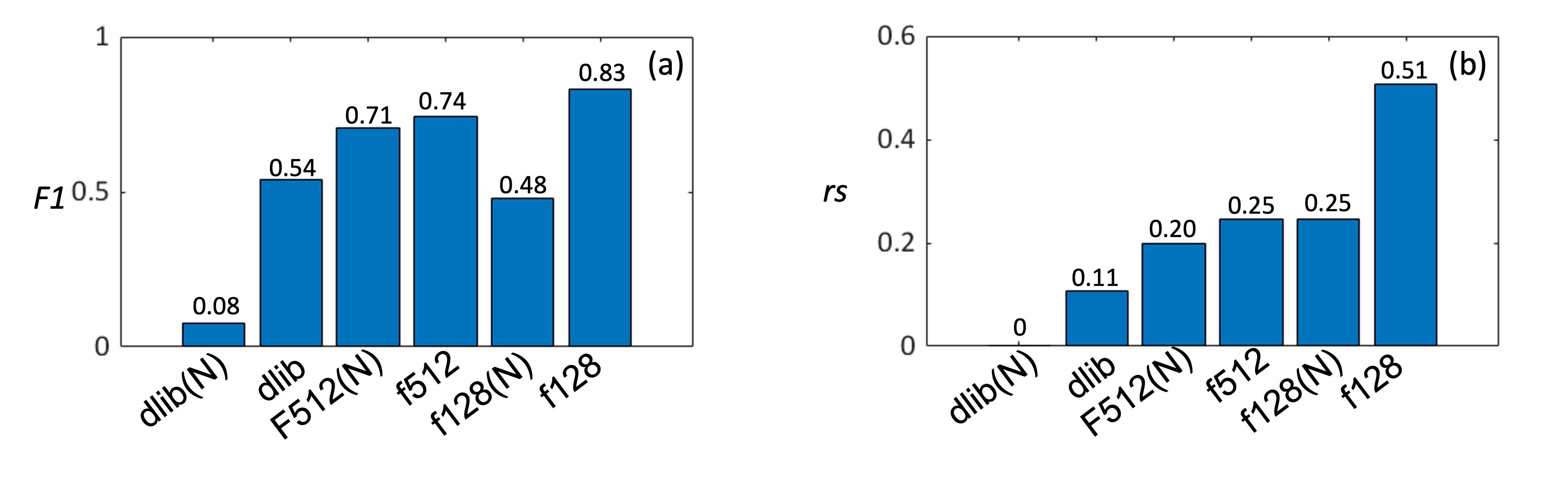} \caption{Results of different encoding methods: (a) $F1$, (b) $rs$.}
\label{fig:encoding}
\end{figure*}
\subsection{Encoding}
\begin{figure*}
\centering 
\includegraphics[width=5.5in]{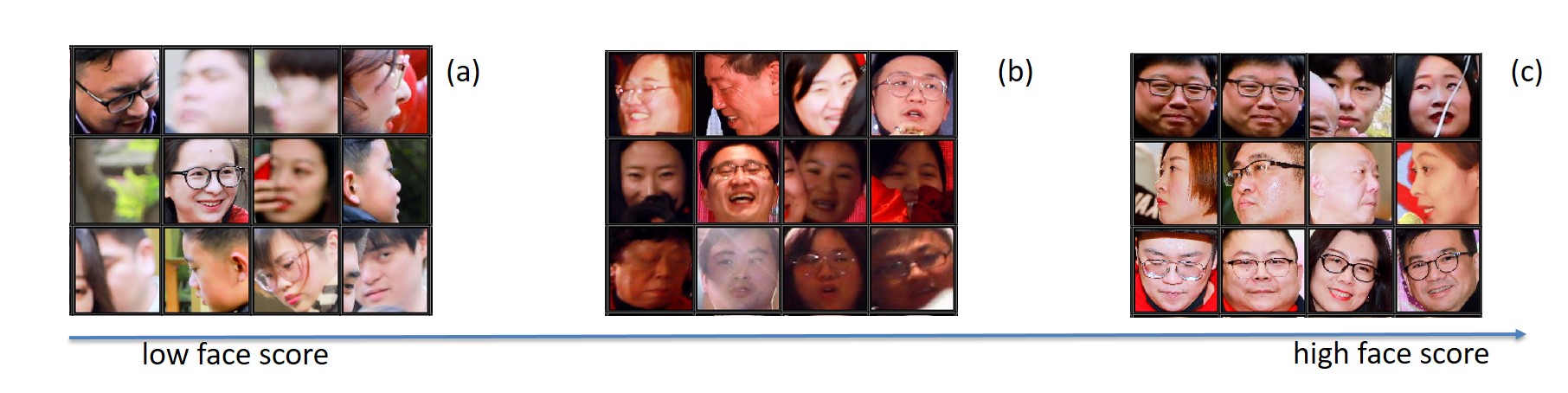}
\caption{Example of faces with different scores: (a) low, (b) median, (c) high.}
\label{fig:faceScore}
\end{figure*}

\noindent
The experiments are conducted using f128. 
As shown in Fig. \ref{fig:faceScore}, the faces in an event image could be blurred, too dark, or too bright, it is desirable for an encoder that can handle these conditions.
Two other different encoding methods are implemented for comparison. 
The first one is dlib, a python library face\_recognition, which encodes faces into a 128-dimensional vector. 
The second one is f512, which is similar to f128 but uses a 512-dimensional vector. 
The comparison is conducted with and without the operations. 
The results are shown in Fig. \ref{fig:encoding}. Note that the result with "(N)" is the one without using any operations. 
The initial clustering algorithm is DBSCAN. 
The result with operations is always better than without the operations. Also, the encoder affects the result of the clustering. 
In general, the encoders with better performance in face recognition perform better. 
With the development of better face encoders, they can be used with the proposed framework for better clustering results.
\section{Discussion}
\label{sec:discussion} 
\noindent 
This section discusses the results of the proposed clustering method. 
It starts with limitations and possible improvements. 
It is followed by the comparisons of the difference between images from social media and events.
\subsection{Limitations and Research Directions}
\noindent 
The method can work with different clustering algorithms \cite{montero2022efficient}, as shown in Fig. \ref{fig:Robustness}, for better results on event images. 
With further development of clustering algorithms, the same framework can be applied to the results of those algorithms for event images, and a better result can be obtained with the steps. 
Although they work well with the events collected, how to optimize the thresholds in face filtering, time grouping, and check duplication are subjects for further investigation. 
As well, how to use the discovered connections is also important. 
A good clustering result can help to detect communities and compute importance of the participants\cite{cheung2021discovering}. 
Hence, it is interesting to investigate how to use the clustering result for it. 
The weights between two participants represent the familiarity value between the two participants which may be considered for a better discovery. 
Another direction is to understand the changing social graphs during the events\cite{chen2012analysis}  for the interactions among the participants during an event, using online clustering algorithms\cite{montero2022efficient}.
\\
\indent
Covid-19 has changed the behaviors of humans. Although humans still need social events, many of the events become online. 
Hence, it is required to understand how connections can be discovered from online events, besides the direct interactions, such as, they are messaging each other online. 
Even for offline events, participants would wear a mask that the current face encoders do not work well, and which creates more difficulties in clustering faces. 
Hence, other visual cues, such as their hairstyle and clothing \cite{matzen2017streetstyle}, are more important for identifying a person in an event image. 
The second direction is using video and audio analytics\cite{lv2018storyrolenet, sharma2020clustering} would help to discover their connections. 
Another direction is using other methods, such as YOLO \cite{tian2019apple}, to detect the whole person for recognition and clustering, and the social connection\cite{wu2010improving} to improve the clustering, can be applied.
\subsection{Difference between images from social media and events}
\noindent 
Images from social media and events differ in many ways. 
First, images shared on social media are selected to contain only information related to the sharer. 
Hence, faces that are on the shared images are likely connected to the sharer. 
The quality of the images would not be too bad, as the sharers would remove those that are not well taken. 
Also, images on social media are more diverse, containing information such as interests. 
On the other hand, images from events are not selectively dropped, and hence, more connections can be discovered from event images. 
With the increasing privacy concerns, it is getting harder to obtain user shared images from social media to obtain connections\cite{cheung2015connection}. 
Event images are more accessible sources to obtain user connections.

\section{Conclusion}
\label{sec:conclusion}
\noindent  
This paper proposes a framework for discovering connections from event images and a clustering method that can enhance any existing clustering algorithms for face clustering on event images. 
It utilizes information from event images such as co-occurrence and time to enhance the initial result from a clustering algorithm. 
Based on the scraped real event data from an online photo live platform with over 40000 faces from 17 events with over 3000 participants, experiments are conducted to prove the robustness of the system with the clustering framework. 
It is proven that the connection can be well discovered with 80\% in F1 score for the discovery of connections and the important people on the social graph. 
Future directions and limitations are also discussed. With the rapid growth of events, there is a great need for connection discovery from images. 
The social graph generated by the proposed algorithm may create a long-term impact on the community and enable many applications that were only available with online data.
\noindent 
\section*{Acknowledgment}
\noindent This work was supported by Socialface HK Limited.
\bibliographystyle{ACM-Reference-Format}
\bibliography{sf}
\newpage
\appendix

\section{Evaluation}
\indent
True positive, ${T^\prime}_p(f_a,f_b\in\ {\bar{\mathbf{U}}}_i\prime)$, and False positive ${F^\prime}_p(f_a,f_b\in\ {\bar{\mathbf{U}}}_{i\prime})$, of 2 faces $f_a$ and $f_b$ are defined as the value of true and false positive for faces $f_a$ and $f_b$ to be in the same cluster, respectively:
\begin{equation}
\label{eq:tp}
{T^\prime}_p(f_a,f_b\in\ {\bar{\mathbf{F}}}_{i\prime})=
\begin{cases}
1\ if\ \ f_a,f_b\in\ \mathbf{F}_i,\\
0 \ \text{otherwise}.
\end{cases}
\end{equation}
\begin{equation}
\label{eq:fp}
{F^\prime}_p(f_a,f_b\in\ {\bar{\mathbf{F}}}_{i\prime})=
\begin{cases}
1\ if\ \ f_a\in\ \mathbf{F}_i,\ f_b\in\ \mathbf{F}_j,\\
0 \ \text{otherwise}.
\end{cases}
\end{equation}
where ${\bar{\mathbf{F}}}_{i\prime}$ is the set of faces belong to cluster $i\prime$, after conducting clustering, and $\mathbf{F}_i$ and $\mathbf{F}_j$ are the set of faces belong to the two participants, $i$ and $j$, labelled in the ground true.
If 2 faces belong to the same participant and the algorithm puts them into the same cluster, ${T^\prime}_p(f_a,f_b\in\ {\bar{\mathbf{F}}}_{i\prime})$ is 1, and it is 0 otherwise.
If 2 faces belong to the 2 different participants and the algorithm puts them into the same cluster, ${F^\prime}_p(f_a,f_b\in\ {\bar{\mathbf{F}}}_i^{\prime(f)})$ is 1, and it is 0 otherwise.
Similarly, ${F^\prime}_N(\ f_a\in\ {\bar{\mathbf{F}}}_{i\prime},\ f_b\in\ {\bar{\mathbf{F}}}_{j\prime})$ is defined as the value of false negative for faces $f_a$ and $f_b$ to be in the different clusters:

\begin{equation}
\label{eq:fn}
{F^\prime}_N(\ f_a\in\ {\bar{\mathbf{F}}}_{i\prime},\ f_b\in\ {\bar{\mathbf{F}}}_{j\prime})=
\begin{cases}
1\ if\ \ f_a,f_b\in\ \mathbf{F}_i,\\
0 \ \text{otherwise}.
\end{cases}
\end{equation}

If 2 faces belong to different participants and the algorithm puts them into the same cluster, ${F^\prime}_N(\ f_a\in\ {\bar{\mathbf{F}}}_{i\prime},\ f_b\in\ {\bar{\mathbf{F}}}_{j\prime})$ is 1, and it is 0 otherwise.

\section{Data Tagging}
\begin{figure*}[h]
\centering 
\includegraphics[width=5.5in]{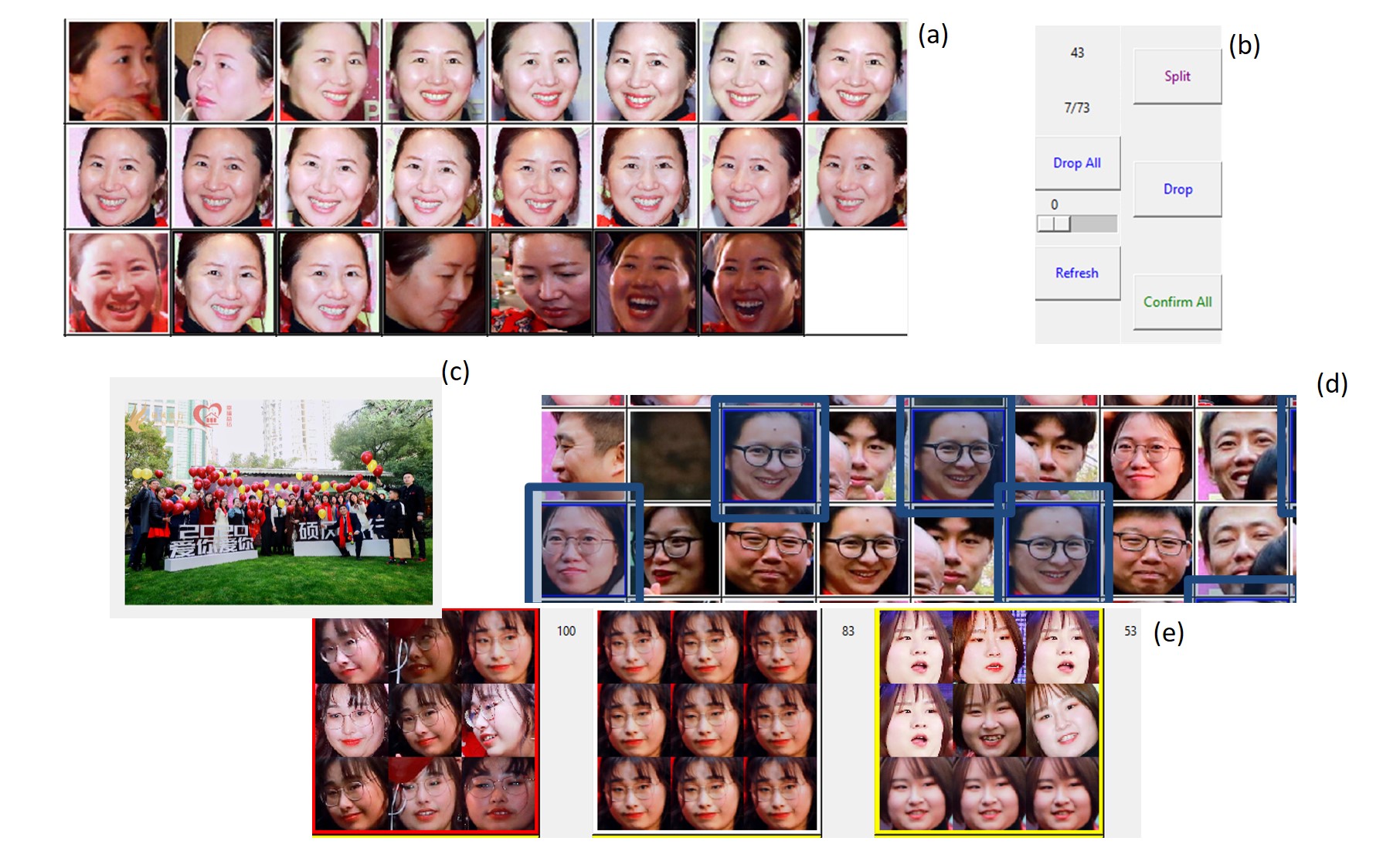} \caption{User interface of the tagging program, (a) show faces of the same clusters for selection, (b) the buttons, (c) show the image when a face is right-clicked, (d) highlight potential faces when a face is left-clicked, (e) show similar clusters for merging.}
\label{fig:TaggingUI}
\end{figure*}

\indent
To cluster faces correctly based on the faces, a program is implemented and shown in Figure \ref{fig:TaggingUI}. 
First, the collected faces are encoded using FaceNet\cite{schroff2015facenet} and then clustered using DBSCAN\cite{rodriguez2014clustering}. 
Clusters are formed, followed by the labeling software that labels faces based on identity manually.
There are five major functionalities in the software. 
The first part shows the faces of the same cluster and some potential faces that are circled in red. 
The user can directly reject or confirm the whole set of faces or select some of the faces to reject or accept, as shown in Figure \ref{fig:TaggingUI}(a). 
The buttons for those operations are shown in Figure \ref{fig:TaggingUI}(b). 
The user can also right-click on the face to check the image that the face is on, as shown in Figure \ref{fig:TaggingUI}(c). 
This way, the user can quickly confirm whether the faces belong to the same person using other cues, such as clothing. 
If there are faces of two persons in the same cluster, the user can select those faces in Figure \ref{fig:TaggingUI}(d) and split the faces of a person to a new cluster. 
To help with the selection, once a face is selected, those similar to the selected faces are labeled in blue, as shown in Figure \ref{fig:TaggingUI}(d). 
Once the faces are confirmed, the mean vector of each cluster is computed, and clusters that are similar to the newly formed cluster are listed, as shown in Figure \ref{fig:TaggingUI}(e). 
The user can click on those clusters and merge them into a cluster. 
Note that some clusters are labeled as yellow, as some of the faces of those clusters have been on the same images with the cluster just confirmed. 
It helps users not to merge clusters that may belong to two different persons accidentally.
\end{document}